\documentclass[11pt]{article}

\usepackage[preprint]{acl}

\usepackage{times}
\usepackage{latexsym}
\usepackage{booktabs}
\usepackage{graphicx}

\usepackage[T1]{fontenc}

\usepackage[utf8]{inputenc}

\usepackage{microtype}

\usepackage{inconsolata}

\usepackage{graphicx}

\usepackage{amsmath}
\usepackage{amssymb}
\usepackage{lipsum}
\usepackage{xurl}
\usepackage[breakable]{tcolorbox}
\usepackage{booktabs}
\usepackage{tabularx}
\usepackage{makecell}
\usepackage[table]{xcolor}
\usepackage{multirow}
\usepackage{graphicx}
\usepackage{placeins}

\newtcolorbox{promptbox}[1][]{
    breakable,
    colback=gray!5,
    colframe=gray!50,
    boxrule=0.5pt,
    arc=3pt,
    left=8pt, right=8pt,
    top=8pt, bottom=8pt,
    fontupper=\small\sffamily,
    title=#1,
    coltitle=black,
    colbacktitle=gray!15,
    toptitle=3pt, bottomtitle=3pt
}

\title{SAEVerbalizer: Generating Explanations for \\ 
Sparse Autoencoder Features via Representation Verbalization}

\author{
\textbf{Weihan Meng}\textsuperscript{1,*},
\textbf{Hongzhu Guo}\textsuperscript{2,*},
\textbf{Yi Jing}\textsuperscript{1},
\textbf{Dewen Liu}\textsuperscript{3},
\\
\textbf{Zijun Yao}\textsuperscript{1},
\textbf{Xiaozhi Wang}\textsuperscript{1},
\textbf{Lei Hou}\textsuperscript{1},
\textbf{Juanzi Li}\textsuperscript{1}
\\
\textsuperscript{1}Tsinghua University,
\textsuperscript{2}Peking University,
\textsuperscript{3}Fudan University
\\
\texttt{mwh24@mails.tsinghua.edu.cn},
\texttt{2400014104@stu.pku.edu.cn}
}

\begin{document}
\maketitle

\begingroup
\renewcommand{\thefootnote}{*}
\footnotetext[0]{Equal contribution.}
\endgroup

\begin{abstract}
Sparse autoencoders (SAEs) are proposed to extract numerous features from large language model (LLM) representations, yet explaining these features still relies primarily on external observation. 
This reliance leads to superficial explanations inferred from observed model behavior and computational inefficiency from collecting such behavioral evidence at scale.
We introduce \textsc{SAEVerbalizer}, a framework that injects SAE decoder directions into an LLM's representations and fine-tunes the LLM's downstream layers to generate natural-language explanations of the injected features. 
Once trained, the resulting verbalizer explains SAE features directly from decoder directions, addressing both limitations.
Our experiments show that the learned verbalization capability generalizes to unseen features, transfers across separately trained SAE dictionaries, and, with a lightweight adapter, extends to SAE features from different LLMs.
Intervention experiments show that injecting multiple directions yields an explanation combining their meanings, while reversing individual directions produces corresponding meaning shifts.
\end{abstract}

\section{Introduction}

Sparse autoencoders~\citep[SAEs; ][]{bricken2023monosemanticity, ICLR2024_1fa1ab11} are widely used~\cite{templeton2024scaling, ICLR2025_42ef3308, gemmascope2_techreport, qwen_scope, zhang2026locate} to interpret internal representations of large language models (LLMs).
SAEs map dense LLM representations into a higher-dimensional sparse feature space, where each dimension is intended to represent an individual interpretable feature. 
However, while SAEs excel at \textbf{\textit{extracting}} features, they fall short of \textbf{\textit{explaining}} them in natural language.

\begin{figure}[t]
    \centering
    \includegraphics[width=\columnwidth]
    {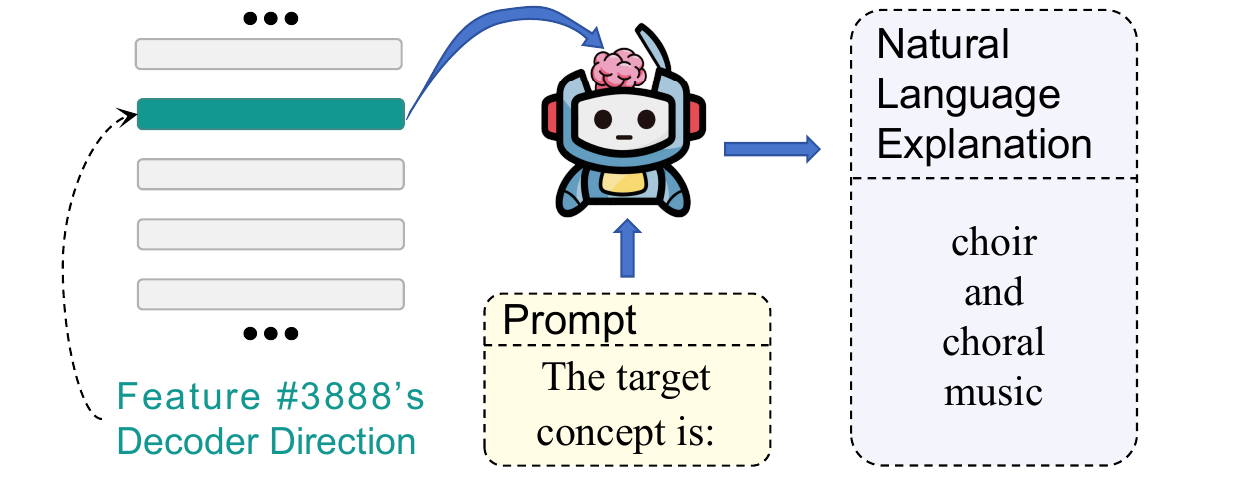}
    \caption{
        Overview of \textsc{SAEVerbalizer}.
        Given an SAE decoder direction and a fixed verbalization prompt,
        the verbalizer generates an explanation of the feature.
        }
    \label{fig:overview}
\end{figure}

Most existing methods explain SAE features through externally observed LLM behavior. 
A prominent approach runs an LLM over a large corpus, identifies each feature's top-activating examples, and prompts an LLM to summarize their shared pattern~\cite{bricken2023monosemanticity,pmlr-v267-paulo25a}. 
However, these methods face two limitations. \textbf{Superficial Explanations.} Without directly examining how a feature is represented inside the LLM, their explanations describe what its activation examples have in common rather than what the feature itself represents.
\textbf{Computational Inefficiency.} Each new SAE requires repeating corpus-scale inference, activation computation, example retrieval, and feature-wise LLM summarization~\cite{pmlr-v267-paulo25a}. These limitations reduce explanation quality and scalability across SAEs.

Recent work has demonstrated LLMs' capability to process internal representations and express their semantic content in natural language~\cite{pmlr-v235-ghandeharioun24a,ICLR2026_7f19b99e,pmlr-v235-chen24ao}. 
Motivated by this capability, we introduce \textsc{SAEVerbalizer}, a framework that shifts SAE feature explanation from observing external behavior to processing internal representations within the LLM.
As shown in Figure~\ref{fig:overview}, it injects an SAE decoder direction—the only feature-specific input—into the LLM's representations and prompts the LLM to generate a natural-language explanation of the corresponding feature.

By explaining SAE features directly from decoder directions, \textsc{SAEVerbalizer} addresses both limitations.
It moves beyond \textbf{superficial explanations} by grounding explanations in information encoded within the LLM's internal representation space, rather than inferring feature meanings from activation texts.
It reduces \textbf{computational inefficiency} by directly verbalizing decoder directions without repeated corpus-scale inference and example retrieval.
Moreover, this capability transfers across SAE dictionaries and, with a lightweight representation-space adapter, extends to SAE features from different LLMs.

Beyond providing a new tool for SAE interpretation, we use this framework to study internal representation verbalization as a trainable capability.
Across multiple LLM scales and SAE layers, this capability generalizes to unseen features and transfers across separately trained SAE dictionaries; lightweight adapters further extend it to SAE features from different LLMs.
Ablations demonstrate gains from additional supervision and robustness to prompt and injection variations.
Intervention analyses illustrate the verbalizer's behavior under joint injection and direction reversal.

Our contributions include: 
(a) proposing \textsc{SAEVerbalizer}, a lightweight approach, to explain unseen SAE features directly from decoder directions; 
(b) providing a practical method for training this capability with feature--explanation supervision; 
and (c) conducting experiments to demonstrate its scalability and generalization across SAE dictionaries and LLMs.
\section{Preliminaries and Related Work}
This section introduces the formal background and terminology needed for our method and situates our approach within the relevant literature.
\subsection{Sparse Autoencoders}
A critical interpretability challenge of LLMs is superposition---an LLM encodes more features than the dimensionality of its representation space permits~\cite{elhage2022superposition}.
To recover interpretable features from such superposed representations, an SAE uses an encoder to map each LLM representation to sparse activations over a highly overcomplete feature dictionary and a decoder to reconstruct the representation from these activations~\cite{bricken2023monosemanticity, ICLR2024_1fa1ab11}.
Formally, given an LLM representation $\mathbf{h}\in\mathbb{R}^{d}$, an SAE computes
\[
\mathbf{z}
=
f\!\left(W_{\mathrm{enc}}\mathbf{h}
+
\mathbf{b}_{\mathrm{enc}}\right),
\quad
\hat{\mathbf{h}}
=
W_{\mathrm{dec}}\mathbf{z}
+
\mathbf{b}_{\mathrm{dec}},
\]
where \(\mathbf{z}\in\mathbb{R}^{m}\) and \(m\) is the SAE dictionary size, typically with \(m\gg d\).
Each coordinate \(z_i\) gives the activation of the \(i\)-th learned feature, whose corresponding decoder column \(\mathbf{w}^{\mathrm{dec}}_i\) defines a direction in the LLM's representation space.
We refer to this direction as the feature's \textbf{decoder direction} and use decoder directions as the interface for representation intervention and verbalization.

\subsection{Automated SAE Feature Interpretation}
Automated SAE feature interpretation primarily follows two paradigms: bottom-up methods infer feature meanings from activation examples, whereas top-down methods locate features associated with predefined concepts.

\paragraph{Bottom-up Interpretation.} 
Bottom-up methods begin with a given SAE feature and infer its meanings from examples that strongly activate it.
Inspired by the automated neuron interpretation approach of \citet{bills2023language}, \citet{bricken2023monosemanticity} apply this paradigm to SAE features by prompting an external LLM to summarize a feature's top-activating examples.
Subsequent work has scaled this paradigm to millions of SAE features~\cite{pmlr-v267-paulo25a} and developed agentic workflows that iteratively propose, test, and refine explanations against activation
evidence~\cite{han-etal-2026-sage}. 
Despite these advances, bottom-up methods remain computationally inefficient, and their explanations can miss features' output effects~\cite{gur-arieh-etal-2025-enhancing}, be overly broad~\cite{NEURIPS2025_5b6ce252}, semantically misaligned~\cite{puri-etal-2025-fade}, or insufficiently discriminative~\cite{mccann2026descriptivecollisionsparseautoencoder}.

\paragraph{Top-down Localization.} 
Top-down methods instead start from a predefined concept and identify SAE features whose activations are associated with it.
Inspired by concept-based interpretation methods such as TCAV~\cite{pmlr-v80-kim18d}, this paradigm has been applied to behavioral concepts~\cite{zhao-etal-2026-denoising}, linguistic phenomena~\cite{jing-etal-2025-lingualens}, and individual languages~\cite{deng-etal-2025-unveiling}.
It supports targeted analysis, but depends on human hypotheses about which concepts to search for and how they should be defined, and therefore does not provide exhaustive explanations of all features in an SAE dictionary.

In contrast, \textsc{SAEVerbalizer} directly maps SAE decoder directions to natural-language explanations, without requiring per-feature analysis of activation examples or a predefined set of concepts.

\subsection{Representation Verbalization and Transferability}

LLMs have the capability to identify concepts injected into their intermediate representations, but this spontaneous introspective ability remains unreliable and sensitive to elicitation~\cite{lindsey2025emergent}.
Prior work shows that LLM computation can be used to decode intermediate representations into natural language~\cite{pmlr-v235-ghandeharioun24a,pmlr-v235-chen24ao,self_explaining_sae_2024}.
Explicit training further enables LLMs to verbalize contextual activations~\cite{frasertaliente2026nla,ICLR2026_7f19b99e}, causal computations~\cite{li2026traininglanguagemodelsexplain}, explain internal features as preference optimization~\cite{he2026saeexplainerinterpretingsaefeatures}, and generalize across representation-interpretation tasks~\cite{DBLP:journals/corr/abs-2512-15674}.
These results together support treating SAE feature verbalization as a learned capability rather than relying on spontaneous introspection.

Extending this capability across LLMs requires aligning their representation spaces.
Such alignment is supported by cross-LLM correspondences in both the local geometry of intermediate representations~\cite{wolfram2025layers} and SAE feature spaces~\cite{lan2025quantifyingfeaturespaceuniversality}.
Related work further shows that mappings can be learned between LLM representation spaces~\cite{NEURIPS2025_4569a868}.
Accordingly, we train adapters to map source-SAE decoder directions into the representation space of the verbalizer's injection layer.
\section{Methodology}

\begin{figure*}[t]
    \centering
    \includegraphics[width=0.98\textwidth]{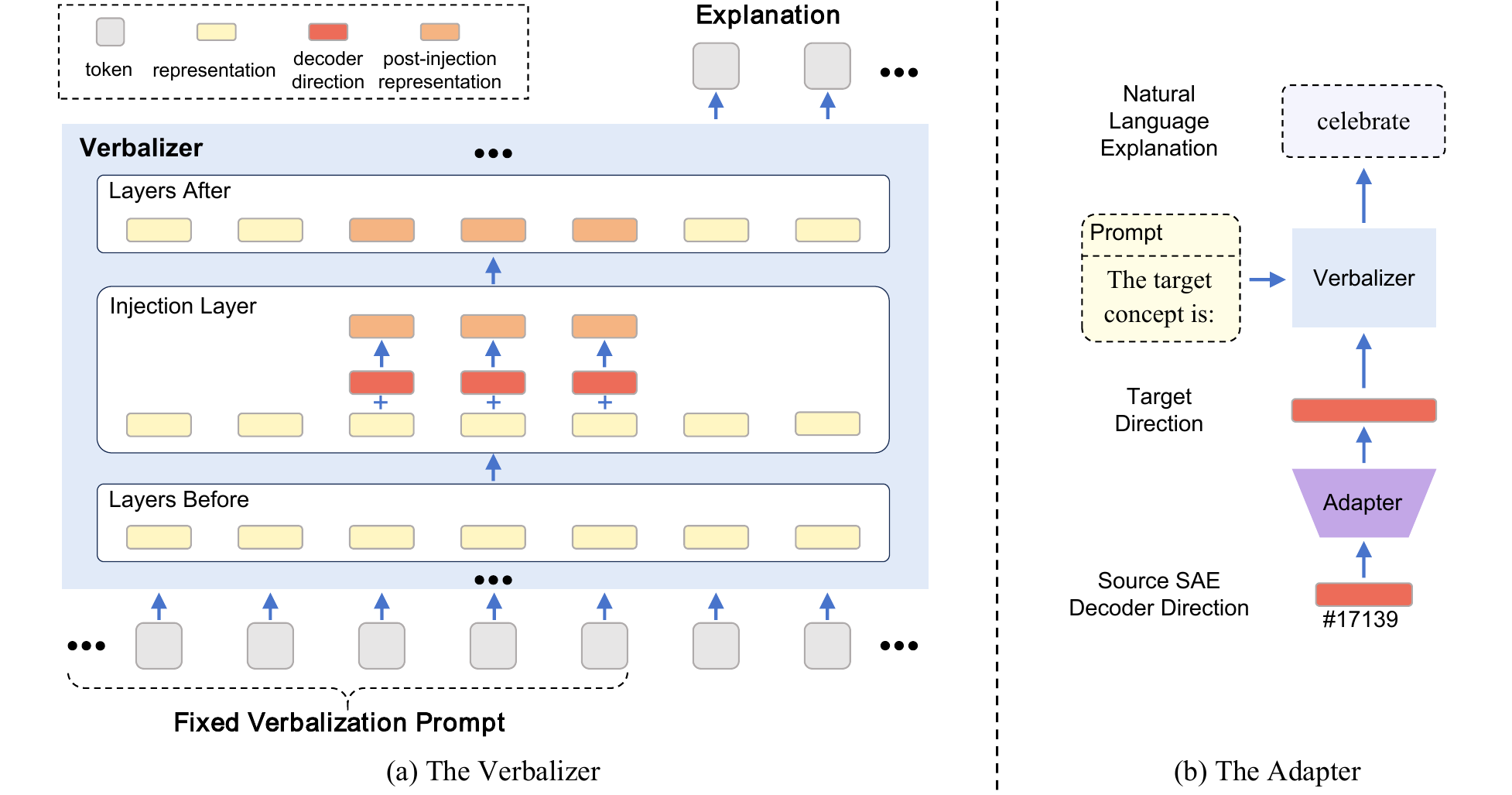}
    \vspace{-0.08in}
    \caption{
        Overall design of \textsc{SAEVerbalizer}.
        (a) \textbf{The Verbalizer.}
        During prompt prefilling, an SAE decoder direction is injected into
        the token representations of a designated injection span at
        \(L_{\mathrm{inj}}\), conditioning explanation generation.
        (b) \textbf{The Adapter.}
        An adapter maps SAE decoder directions in a source LLM's
        representation space into the verbalizer's injection-layer
        representation space.
        }
    \label{fig:method}
\end{figure*}

We fine-tune LLMs to generate natural-language explanations from injected SAE decoder directions, referring to the resulting models as \textit{verbalizers} and to the LLMs used to initialize them as their \textit{verbalizer backbones}.
During prompt prefilling, a decoder direction is injected at the verbalizer's injection layer into the token representations of a designated injection span.
Downstream computation over the post-injection representations conditions subsequent generation, yielding an explanation.
For SAE features from another LLM, an \textit{adapter} maps their decoder directions into the verbalizer's injection-layer representation space.
Figure~\ref{fig:method} shows an overview of \textsc{SAEVerbalizer}.

\subsection{Verbalizer Design}
\label{sec:verbalizer_design}
The verbalizer receives a fixed, feature-agnostic prompt that specifies the verbalization task.
During prompt prefilling, the target decoder direction is injected at layer \(L_{\mathrm{inj}}\) into each token representation in a designated injection span
following the task instruction, serving as the only feature-specific input.
No activation examples or their surrounding contexts are provided to the verbalizer.
The remaining Transformer layers process the post-injection prompt representations during prefilling, conditioning subsequent autoregressive generation to yield an explanation.

Formally, let
\(\mathbf{H}=[\mathbf{h}_{b,s}]\in\mathbb{R}^{B\times S\times D}\)
denote the pre-injection representations over the injection span at layer
\(L_{\mathrm{inj}}\), where \(B\), \(S\), and \(D\) are the batch size, injection-span length, and representation dimension, respectively.
For batch element \(b\), \(f_b\) denotes the target feature and
\(\mathbf{v}_{f_b}\in\mathbb{R}^{D}\) its decoder direction in the verbalizer's injection-layer representation space.

We apply the following norm-matched additive injection independently to each batch element:
\begin{equation}
\mathbf{h}'_{b,s}
=
\mathbf{h}_{b,s}
+
\alpha \bar{n}_b \hat{\mathbf{v}}_{f_b},
\qquad
s=1,\ldots,S,
\end{equation}
where
\(\hat{\mathbf{v}}_{f_b}
=
\mathbf{v}_{f_b}/(\lVert\mathbf{v}_{f_b}\rVert_2+\epsilon)\)
is the normalized decoder direction
(\(\epsilon>0\) ensures numerical stability), and
\(\bar{n}_b = \frac{1}{S}\sum_{s=1}^{S}\lVert\mathbf{h}_{b,s}\rVert_2\)
is the mean pre-injection representation norm over the injection span.
This normalization removes the variation in direction norms across features, while \(\bar{n}_b\) scales the injected direction to the local representation magnitude.
\(\alpha\) is used to control its relative strength.

Once trained, the verbalizer can be applied to unseen features without additional supervision or fine-tuning.
For each feature, it processes the short fixed task prompt with the corresponding decoder direction injected and generates an explanation.

\subsection{Adapter Design}
\label{sec:adapter_design}
The verbalizer processes decoder directions in its injection-layer representation space, while directions from an SAE trained on another LLM lie in a different space.
To reuse the verbalizer across LLMs, we introduce a lightweight adapter that maps a selected source-layer representation space to the verbalizer's
injection-layer representation space.
Motivated by cross-LLM correspondences between representations at similar
depths~\cite{wolfram2025layers}, we select source and target layers at comparable relative depths within their respective LLMs.

Specifically, we parameterize the adapter as a single affine layer:
\begin{equation}
A(\mathbf{x})
=
W\mathbf{x}+\mathbf{b},
\quad
W\in\mathbb{R}^{d_t\times d_s},
\
\mathbf{b}\in\mathbb{R}^{d_t},
\end{equation}
where $d_s$ and $d_t$ are the representation dimensions of the selected source-LLM layer and the verbalizer injection layer, respectively.

Although trained to align representations from the two layers, the adapter is used to map source-SAE decoder directions at inference.
Because decoder directions enter the verbalizer through additive injection, we map each direction according to the change it induces in the adapter output.
For a source-layer representation \(\mathbf{h}^{(s)}\) and a source-SAE decoder direction \(\mathbf{v}^{(s)}_f\), this change, denoted
\(\tilde{\mathbf{v}}^{(t)}_f\), is
\(A(\mathbf{h}^{(s)}+\mathbf{v}^{(s)}_f)-A(\mathbf{h}^{(s)})
= W\mathbf{v}^{(s)}_f\).
The affine bias cancels in this difference and does not contribute to the mapped direction.

The mapped direction is then supplied to the verbalizer through the same normalization, injection, and generation interface defined above.
Once the adapter has been trained, the existing verbalizer can be applied to SAE decoder directions from the source LLM without collecting source-specific feature--explanation supervision or fine-tuning a separate verbalizer for that LLM.

\subsection{Training Method}
\label{sec:training}
We fine-tune the verbalizer on high-quality feature--explanation pairs and train the adapter on aligned representation pairs from source and target layers.

\subsubsection{Verbalizer Fine-Tuning}
\label{sec:verbalizer_training}
We fine-tune the verbalizer by injecting each feature's decoder direction and using its paired explanation as the generation target, while updating only the components downstream of the injection layer.

\paragraph{Feature--Explanation Supervision.}
We obtain candidate feature--explanation pairs from Neuronpedia~\cite{lin2023neuronpedia}, where explanations are inferred from a feature's top-activating text examples.
Because these observational explanations vary substantially in reliability (Appendix~\ref{app:data_preliminary}), we use the LLM filtering judge to score three properties: the coherence of the pattern shared across the examples, the
specificity of the explanation, and how consistently the examples support it.
For training, we retain candidates whose scores meet the training qualification standard.
Appendix~\ref{app:data_filtering} provides the full filtering procedure, qualification standards, and judge prompts.

\paragraph{Training Objective.}
For each training pair \((f,y_f)\), we inject the corresponding decoder direction \(\mathbf{v}_f\) using the interface defined in Section~\ref{sec:verbalizer_design} and optimize the verbalizer to generate \(y_f\). 
We use the standard causal language modeling objective under teacher forcing, computing token-level cross-entropy only over the explanation tokens while masking the prompt tokens from the loss.

\paragraph{Partial Fine-Tuning.}
We partition the Transformer layers at \(L_{\mathrm{inj}}\) as follows:
\[
l \leq L_{\mathrm{inj}}
\quad \text{frozen},
\qquad
l > L_{\mathrm{inj}}
\quad \text{trainable}.
\]
Because injection occurs at the output of \(L_{\mathrm{inj}}\), only downstream components process the post-injection representations; freezing the upstream layers also preserves the representation space in which the SAE decoder directions are defined.
We additionally freeze the embedding layer and keep the SAE and its decoder directions fixed, while fine-tuning the final normalization layer and language modeling head.

\subsubsection{Adapter Training}
\label{sec:adapter_training}
We train the adapter to map source-LLM representations into the verbalizer's representation space using aligned hidden representations from unlabeled text, while keeping both LLMs frozen.
\paragraph{Training Data.}
The adapter is trained on ordinary unlabeled text and requires no feature--explanation supervision. 
We pass the same token sequence through the frozen source LLM and verbalizer backbone. 
For each non-padding token position \(i\), we pair the representation \(\mathbf{x}_i\in\mathbb{R}^{d_s}\) from the selected source-LLM layer with the representation \(\mathbf{y}_i\in\mathbb{R}^{d_t}\) from the verbalizer injection layer, both produced from the same token and preceding context. 
For the LLM pairs considered in this work, a shared tokenizer provides token-level alignment.

\paragraph{Training Objective.}
For a batch containing N aligned token positions, we train the adapter to reconstruct the corresponding verbalizer representations:
\begin{equation}
    \mathcal{L}_{\mathrm{adapter}}
    =
    \frac{1}{N d_t}
    \sum_{i=1}^{N}
    \left\|
        W\mathbf{x}_i+\mathbf{b}-\mathbf{y}_i
    \right\|_2^2.
\end{equation}
Only the adapter parameters \(W\) and \(\mathbf{b}\) are updated, while the source LLM and verbalizer backbone remain frozen.

\section{Experiment}
\label{sec:experiments}
Our experiments evaluate the feasibility of direct SAE feature
verbalization from decoder directions and further analyze the verbalization capability.

\subsection{Experiment Setup}
\label{sec:experiment_setup}
This subsection specifies the verbalizer and adapter configurations, test sets, and evaluation protocol used throughout the experiments.
\paragraph{Verbalizer Configurations.}
For the main experiments, we fine-tune \nolinkurl{gemma-3-1b-it}, \nolinkurl{gemma-3-4b-it}, and \nolinkurl{gemma-3-27b-it}~\cite{gemmateam2025gemma3technicalreport} as verbalizers.
We use width-262k, medium-sparsity Gemma Scope~2 SAEs~\cite{gemmascope2_techreport} trained on post-layer residual-stream representations.
Each verbalizer corresponds to one selected SAE and injects its decoder directions at the matching layer.
Across the 1B, 4B, and 27B backbones, the selected SAE layers are \(\{7,13,17,22\}\), \(\{9,17,22,29\}\), and \(\{16,31,40,53\}\), respectively.

Training uses nested sets of 12k, 24k, and 48k qualified feature--explanation pairs for each SAE.
We use 12k as the common comparison point and report results at 24k and 48k where sufficient qualified data are available.

We use the \texttt{27B-L16} configuration with 48k
feature--explanation pairs as the default for analysis experiments and refer to the resulting model as the \textit{default verbalizer}.
Full implementation settings, including backbone-specific learning rates, are provided in Appendix~\ref{app:verbalizer_implementation_details}.

\paragraph{Adapter Configurations.}

For cross-LLM transfer, we train adapters from \texttt{1B-L7} and \texttt{4B-L9} into the default 27B verbalizer's layer-16 representation space.
The two source layers and the target layer lie at approximately one-quarter of their respective LLM depths.
Additional implementation details are provided in Appendix~\ref{app:adapter_details}.

\paragraph{Test Sets.}
\label{sec:test_sets}
Using the LLM-based filtering procedure, we construct three disjoint test sets for each SAE evaluated.
The \textit{Global Train-Standard} (GTS) set contains 1,000 globally sampled features satisfying the training qualification standard.
The \textit{Low-Index Gold} (LIG) and \textit{Global Gold} (GG) sets contain 200 low-index and 1,000 globally sampled features, respectively, satisfying a stricter gold qualification standard.
GTS and GG differ in qualification strictness, whereas LIG and GG differ in index distribution; the latter comparison probes features prioritized by Gemma Scope~2's Matryoshka reconstruction objective~\cite{gemmascope2_techreport,pmlr-v267-bussmann25a}.

Within each SAE dictionary used for verbalizer training, the training and test sets are disjoint.
Further details of the test-set construction procedures are provided in Appendix~\ref{app:data_filtering}.

\paragraph{Evaluation Protocol.}
We evaluate generated explanations using Reference Agreement (RA).
For each test feature, the LLM evaluation judge compares the generated explanation with its Neuronpedia reference.
RA is the proportion of test features for which the two explanations are judged to agree.

Quality filtering supports using RA as a scalable proxy for verbalization quality.
However, because the references are inferred from activation examples rather than established ground truth, RA does not establish the absolute correctness of generated explanations.
The judge prompt and judge decoding settings are provided in Appendix~\ref{app:evaluation_protocol}.

\subsection{Experiment Results}
\label{sec:main_results}
\begin{table}[htbp]
\centering
\scalebox{0.89}{
\begin{tabular*}{\columnwidth}{
cccrrr
}
\toprule
\textbf{LLM} &
\textbf{Layer} &
\textbf{\#Train} &
\textbf{GTS} &
\textbf{LIG} &
\textbf{GG} \\
\midrule
\multirow{8}{*}{
\rotatebox{90}{\texttt{gemma-3-1b-it}}
}
& \multirow{2}{*}{7}
& 12k & $19.6$ & $33.0$ & $14.6$ \\
& & 24k & $17.6$ & $35.5$ & $14.7$ \\
\cmidrule(l{4pt}r{2pt}){2-6}

& 13
& 12k & $18.7$ & $27.0$ & $17.1$ \\
\cmidrule(l{4pt}r{2pt}){2-6}

& \multirow{2}{*}{17}
& 12k & $8.9$ & $14.5$ & $8.6$ \\
& & 24k & $8.2$ & $17.5$ & $8.0$ \\
\cmidrule(l{4pt}r{2pt}){2-6}

& \multirow{3}{*}{22}
& 12k & $3.9$ & $7.0$ & $4.2$ \\
& & 24k & $5.1$ & $11.0$ & $6.1$ \\
& & 48k & $5.5$ & $10.0$ & $6.7$ \\

\midrule

\multirow{8}{*}{
\rotatebox{90}{
\texttt{gemma-3-4b-it}
}}
& \multirow{2}{*}{9}
& 12k & $32.9$ & $51.5$ & $35.2$ \\
& & 24k & $37.6$ & $56.0$ & $39.4$ \\
\cmidrule(l{4pt}r{2pt}){2-6}

& 17
& 12k & $29.4$ & $44.5$ & $30.4$ \\
\cmidrule(l{4pt}r{2pt}){2-6}

& \multirow{2}{*}{22}
& 12k & $21.9$ & $27.5$ & $20.6$ \\
& & 24k & $27.2$ & $32.0$ & $24.4$ \\
\cmidrule(l{4pt}r{2pt}){2-6}

& \multirow{3}{*}{29}
& 12k & $11.2$ & $15.0$ & $10.0$ \\
& & 24k & $11.2$ & $11.5$ & $9.2$ \\
& & 48k & $14.1$ & $16.5$ & $13.4$ \\

\midrule

\multirow{8}{*}{
\rotatebox{90}{\texttt{gemma-3-27b-it}}
}
& \multirow{3}{*}{16}
& 12k & $48.1$ & $80.5$ & $49.2$ \\
& & 24k & $51.0$ & $78.5$ & $50.6$ \\
& & 48k & $52.3$ & $80.5$ & $56.1$ \\
\cmidrule(l{4pt}r{2pt}){2-6}

& 31
& 12k & $39.4$ & $62.0$ & $39.6$ \\
\cmidrule(l{4pt}r{2pt}){2-6}

& \multirow{2}{*}{40}
& 12k & $40.3$ & $55.5$ & $37.1$ \\
& & 24k & $46.9$ & $60.0$ & $43.9$ \\
\cmidrule(l{4pt}r{2pt}){2-6}

& \multirow{2}{*}{53}
& 12k & $24.5$ & $46.0$ & $20.2$ \\
& & 24k & $26.7$ & $51.5$ & $24.5$ \\

\bottomrule
\end{tabular*}
}
\caption{
RA (\%) across verbalizer backbones, SAE layers, and available
training-set sizes.
}
\label{tab:main_verbalizer_results}
\end{table}

Table~\ref{tab:main_verbalizer_results} reports RA across backbone scales, SAE layers, and training-set sizes.
The results show that verbalizers fine-tuned with feature--explanation supervision generalize to unseen features across all evaluated backbone--layer configurations.
The best-performing configuration, \texttt{27B-L16} with 48k pairs, achieves RA scores of 52.3\%, 80.5\%, and 56.1\% on GTS, LIG, and GG, respectively.
At the common training-set size of 12k, RA generally increases with backbone scale and tends to be higher at earlier layers within each backbone.
The backbone-scale trend is consistent with differences in model capacity, and the layer trend is consistent with the amount of downstream computation available after injection.
Because each configuration uses a distinct SAE, both trends may also partly reflect variation across SAE feature distributions.

\subsection{Analysis Experiments}
\label{sec:analysis_experiments}

We next empirically analyze the verbalizer from three perspectives: transfer across SAE dictionaries and LLMs, sensitivity to key design choices, and qualitative behavior under individual, joint, and sign-reversed feature injection.

\subsubsection{Transferability}
\label{sec:verbalizer_transfer}
\textsc{SAEVerbalizer} exhibits transferability across SAE dictionaries and LLMs through two distinct mechanisms.
Within a shared representation space, the verbalizer can be reused directly across SAE dictionaries, whereas lightweight adapters enable transfer across LLM-specific representation spaces.
We evaluate transferability in both settings.

\paragraph{Transfer Across SAE Dictionaries.}
We directly apply the default verbalizer to decoder directions from an unseen width-65k Gemma Scope 2 SAE defined on the same LLM and layer.
We construct its test sets using the same procedure as for the
width-262k SAE.
Because the two SAEs share the same representation space, transfer requires neither an adapter nor additional feature--explanation supervision.
Table~\ref{tab:cross_sae_transfer} shows that the default verbalizer achieves substantial agreement on the unseen width-65k SAE.
This result indicates that the learned verbalization capability is not tied to the SAE dictionary used for training and can be reused directly across SAE dictionaries in the same representation space.

\begin{table}[htbp]
\centering
\scalebox{0.89}{
\begin{tabular}{lrrr}
\toprule
\textbf{Target SAE} & \textbf{GTS} & \textbf{LIG} & \textbf{GG} \\
\midrule
Width-262k & $52.3$ & $80.5$ & $56.1$ \\
Width-65k  & $64.4$ & $56.5$ & $65.9$ \\
\bottomrule
\end{tabular}
}
\caption{
RA (\%) of the default verbalizer on width-262k and width-65k SAEs at the same LLM layer; supervision uses only the width-262k SAE.
}
\label{tab:cross_sae_transfer}
\end{table}

\paragraph{Transfer Across LLMs.}
Using separately trained adapters, we map decoder directions from the width-262k SAEs at \texttt{1B-L7} and \texttt{4B-L9} into the layer-16 representation space of the default 27B verbalizer.
Each adapter-based verbalization system is compared with the
corresponding native verbalizer fine-tuned on 24k feature--explanation pairs, using the same test features from the source SAE.
Table~\ref{tab:cross_model_adaptation} shows that the
verbalization capability can be transferred across LLM-specific representation spaces through lightweight adapters, without source-specific feature--explanation supervision or further fine-tuning of the verbalizer.
For \texttt{1B-L7}, adapter-based transfer improves performance over the native verbalizer on all three test sets, demonstrating that cross-LLM adaptation can leverage the stronger verbalization capability of the 27B verbalizer for SAE features from a smaller LLM.
For \texttt{4B-L9}, whose native verbalizer already achieves higher performance, transfer does not yield additional gains.

\begin{table}[htbp]
\centering
\setlength{\tabcolsep}{6pt}
\scalebox{0.89}{
\begin{tabular}{llrrr}
\toprule
\textbf{Source} &
\textbf{System} & \textbf{GTS} & \textbf{LIG} & \textbf{GG} \\
\midrule
\multirow{2}{*}{\texttt{1B-L7}}
& Native (24k)
& $17.6$ & $35.5$ & $14.7$ \\
& Adapter + 27B
& $18.2$ & $39.0$ & $17.4$ \\
\midrule
\multirow{2}{*}{\texttt{4B-L9}}
& Native (24k)
& $37.6$ & $56.0$ & $39.4$ \\
& Adapter + 27B
& $32.8$ & $49.0$ & $32.1$ \\
\bottomrule
\end{tabular}
}
\caption{
RA (\%) for native and adapter-based verbalization of SAE features from the 1B and 4B LLMs.
}
\label{tab:cross_model_adaptation}
\end{table}

\subsubsection{Ablation Study}
\label{sec:ablations}

Starting from the default configuration, we vary one factor at a time to examine the effects of supervision size, prompt, injection span, injection mode, and injection strength.
For prompt and injection ablations, each variant is used consistently during fine-tuning and inference.

\paragraph{Scaling with Feature--Explanation Supervision.}
We fine-tune the \texttt{27B-L16} verbalizer on nested subsets of 1.5k--48k feature--explanation pairs, using the corresponding verbalizer backbone as a zero-supervision reference.
Table~\ref{tab:data_size_ablation} shows a large gain over the verbalizer backbone after fine-tuning on 1.5k pairs.
Further supervision improves RA on GTS and GG, while performance on LIG reaches a high level early and remains relatively stable.
Because the training pairs are globally sampled and randomly ordered, LIG's earlier saturation is unlikely to result from preferential exposure to low-index features and instead suggests that they require less supervision.
These results show that limited feature--explanation supervision is sufficient to acquire verbalization capability, while additional data primarily improves performance across the broader SAE feature distribution.

\begin{table}[hbtp]
\centering
\scalebox{0.89}{
\begin{tabular}{
l r r r
}
\toprule
\textbf{\#Train} &
\multicolumn{1}{c}{\textbf{GTS}} &
\multicolumn{1}{c}{\textbf{LIG}} &
\multicolumn{1}{c}{\textbf{GG}} \\
\midrule
Backbone & $1.6$  & $2.5$  & $1.2$ \\
1.5k      & $36.4$ & $74.0$ & $41.3$ \\
3k        & $38.4$ & $76.0$ & $40.7$ \\
6k        & $41.8$ & $76.0$ & $44.2$ \\
12k       & $48.1$ & $80.5$ & $49.2$ \\
24k       & $51.0$ & $78.5$ & $50.6$ \\
48k       & $52.3$ & $80.5$ & $56.1$ \\
\bottomrule
\end{tabular}
}
\caption{
RA (\%) across supervision sizes, with the verbalizer backbone as a zero-supervision reference.
}
\label{tab:data_size_ablation}
\end{table}

\paragraph{Prompt and Injection Design.} 
Semantically similar prompts, nearby prompt-aligned injection spans, and additive versus interpolative injection produce only minor differences in performance, with full results reported in Appendix~\ref{app:prompt_injection_robustness}.
As shown in Table~\ref{tab:injection_strength}, performance is also broadly stable across a wide range of injection strengths, but degrades substantially when the injected direction becomes too weak. 
Together, these results show that the verbalizer is robust to reasonable variations in its prompt and injection interface, while requiring sufficiently strong feature injection.

\begin{table}[hbtp]
\centering
\scalebox{0.89}{
\begin{tabular}{
r r r r
}
\toprule
\multicolumn{1}{c}{\(\boldsymbol{\alpha}\)} &
\multicolumn{1}{c}{\textbf{GTS}} &
\multicolumn{1}{c}{\textbf{LIG}} &
\multicolumn{1}{c}{\textbf{GG}} \\
\midrule
$0.01$ & $18.9$ & $44.5$ & $16.2$ \\
$0.10$ & $50.6$ & $78.5$ & $54.2$ \\
$0.20$ & $52.3$ & $80.5$ & $56.1$ \\
$0.30$ & $52.8$ & $81.0$ & $54.8$ \\
$0.60$ & $53.6$ & $81.0$ & $54.2$ \\
$1.00$ & $52.7$ & $80.5$ & $53.4$ \\
\bottomrule
\end{tabular}
}
\caption{
RA (\%) across additive injection strengths.
The default strength is \(\alpha=0.2\).
}
\label{tab:injection_strength}
\end{table}

\subsubsection{Case Study}
\label{sec:case_studies}
Using the default verbalizer, we complement the quantitative evaluation with three qualitative analyses: comparisons with Neuronpedia explanations, joint feature injection, and direction reversal.
\begin{table*}[t]
\centering
\small
\setlength{\tabcolsep}{4pt}
\begin{tabularx}{\textwidth}{
>{\raggedright\arraybackslash}p{0.13\textwidth}
>{\raggedright\arraybackslash}p{0.07\textwidth}
>{\raggedright\arraybackslash}p{0.18\textwidth}
>{\raggedright\arraybackslash}p{0.13\textwidth}
X
}
\toprule
\textbf{Pattern} &
\textbf{Feature} &
\textbf{Neuronpedia} &
\textbf{Verbalizer} &
\textbf{Activation Evidence} \\
\midrule

\textbf{Recurring lexical form} &
\#14949 &
lasting, enduring, eternal states &
forever &
The maximally activated token is consistently \textit{forever} across
varied contexts. \\

\arrayrulecolor{black!20}
\specialrule{0.4pt}{2pt}{2pt}
\arrayrulecolor{black}

\textbf{Localized structural pattern} &
\#115968 &
Gu followed by letters or syllables &
gu- prefix &
The feature activates on word-initial \textit{gu} in otherwise unrelated
words, such as \textit{guarana}, \textit{gucci}, and
\textit{guillaume}, among many others. \\

\arrayrulecolor{black!20}
\specialrule{0.4pt}{2pt}{2pt}
\arrayrulecolor{black}

\textbf{Contextual semantic abstraction} &
\#85263 &
energetic social activities and atmospheres &
lively atmosphere &
The maximally activated token is consistently \textit{lively}, appearing
in varied contexts such as debates, music, places,
and atmospheres. \\

\bottomrule
\end{tabularx}

\caption{
Selected cases in which verbalizer-generated and Neuronpedia
explanations differ in focus or granularity.
}
\label{tab:qualitative_explanation_comparison}
\end{table*}
\paragraph{Qualitative Comparison with Neuronpedia Explanations.} 
We inspect the top-activating text examples for features from the quality-filtered test sets and an additional random test set sampled globally from the full SAE dictionary.
Table~\ref{tab:qualitative_explanation_comparison} presents selected cases in which the verbalizer-generated and Neuronpedia explanations differ in focus or granularity.
In these cases, the verbalizer identifies localized lexical or structural patterns, or different semantic abstractions, illustrating the potential of direct verbalization of decoder directions to complement explanations inferred from top-activating text examples.
Additional examples are provided in
Appendix Table~\ref{tab:additional_qualitative_comparisons}.

\paragraph{Compositionality under Joint Feature Injection.}
We jointly inject two SAE decoder directions with equal coefficients while keeping the total injection strength the same as in single-feature injection, an operation related to prior work on composing representation-space interventions~\cite{han-etal-2024-word,scalena-etal-2024-multi}.
As shown in Table~\ref{tab:feature_composition_cases}, the resulting verbalizations preserve information from both constituent features, either by combining them into a natural joint concept or by expressing a meaningful relation between them. 
Additional examples are provided in Appendix Table~\ref{tab:additional_feature_composition}.

\begin{table}[htbp]
\centering
\small
\setlength{\tabcolsep}{3pt}
\begin{tabular}{
l
>{\raggedright\arraybackslash}p{0.43\columnwidth}
>{\raggedright\arraybackslash}m{0.31\columnwidth}
}
\toprule
\textbf{Feature} &
\textbf{Standard} &
\textbf{Joint} \\
\midrule

\#10040
& exhaustion and fatigue
& \multirow{2}{=}{deadlines and exhaustion} \\
\#53806
& deadlines and timeframes
& \\

\arrayrulecolor{black!20}
\specialrule{0.4pt}{2pt}{2pt}
\arrayrulecolor{black}

\#2862
& river names and descriptions
& \multirow{2}{=}{\raisebox{-1.5ex}{river overflow}} \\
\#69595
& overflowing
& \\

\arrayrulecolor{black!20}
\specialrule{0.4pt}{2pt}{2pt}
\arrayrulecolor{black}

\#15187
& love and enthusiasm
& \multirow{2}{=}{love of coffee} \\
\#1586
& coffee and its contexts
& \\

\bottomrule
\end{tabular}

\caption{
Selected examples of joint feature injection.
The two directions receive equal coefficients summing to the default
single-feature strength, \(\alpha=0.2\).
}
\label{tab:feature_composition_cases}
\end{table}

\paragraph{Meaning Shifts under Direction Reversal.}
We replace the standard injection coefficient \(\alpha\) with \(-\alpha\) while keeping all other settings fixed.
Standard SAEs do not encode opposite concepts as opposite signs of a single feature~\cite{ICLR2026_527360ca}; thus, reversing a decoder direction need not produce the opposite concept.
Table~\ref{tab:sign_reversal_cases} shows that the sign-reversed verbalizations remain semantically related to their standard counterparts.
Moreover, features with related standard verbalizations undergo corresponding shifts after reversal despite low cosine similarities between their decoder directions.
Together, these cases suggest that the verbalizer captures how meanings vary with decoder-direction sign, rather than assigning each feature a fixed explanation.

\begin{table}[htbp]
\centering
\small
\setlength{\tabcolsep}{3pt}
\renewcommand{\arraystretch}{1.12}

\begin{tabular}{
l
>{\raggedright\arraybackslash}m{0.37\columnwidth}
>{\raggedright\arraybackslash}m{0.29\columnwidth}
c
}
\toprule
\textbf{Feature} &
\textbf{Standard} &
\textbf{Sign-Reversed} &
\textbf{Cos.} \\
\midrule

\#40105
& lessons learned
& lesson plan
& \multirow{2}{*}{$0.090$} \\
\#111800
& lessons
& lesson plan
& \\

\arrayrulecolor{black!20}
\specialrule{0.4pt}{2pt}{2pt}
\arrayrulecolor{black}

\#7905
& recent events and time
& dates and years
& \multirow{2}{*}{$0.178$} \\
\#154632
& recent events
& dates and years
& \\

\arrayrulecolor{black!20}
\specialrule{0.4pt}{2pt}{2pt}
\arrayrulecolor{black}

\#6187
& accessing information or services
& to do
& \multicolumn{1}{c}{
    \multirow{2}{*}[-1.2ex]{$0.002$}
  } \\
\#8988
& more information
& how to do something
& \\

\bottomrule
\end{tabular}
\caption{
Selected feature pairs under standard (\(\alpha\)) and sign-reversed
(\(-\alpha\)) injection.
Cos. is the within-pair cosine similarity of their decoder directions.
}
\label{tab:sign_reversal_cases}
\end{table}

\section{Discussion on Similarity with Concurrent Work}
\textsc{SAEVerbalizer} shares substantial methodological similarities with a concurrent work SAEExplainer~\cite{he2026saeexplainerinterpretingsaefeatures}, as both approaches inject representations into LLMs to generate explanations for SAE features. Our work was initiated independently in November 2025, and the methodological similarities between the two works arose independently. The two approaches also differ in several important aspects. \textsc{SAEVerbalizer} does not employ iterative DPO to train the LLM for feature explanation. In addition, we investigate the transferability of the learned verbalization capability across different SAE dictionaries and LLMs. 

A direct empirical comparison with SAEExplainer is difficult due to differences in experimental settings and the availability of released artifacts. Specifically, our experiments use the Gemma 3 and Gemma Scope 2 families of LLMs and SAEs, respectively, which differ from those used in SAEExplainer. Moreover, to the best of our knowledge, their private trained checkpoints and data have not been made public. We therefore regretfully cannot offer a direct comparison, and encourage readers to consider the methodological and experimental differences between the two approaches when interpreting their respective results.

\section{Conclusion}

We introduced \textsc{SAEVerbalizer}, a framework that fine-tunes an LLM's downstream layers to generate natural-language explanations of SAE features from injected decoder directions.
Experiments show that the learned verbalization capability generalizes to unseen features, transfers to unseen SAE dictionaries without further fine-tuning, and extends to SAE features from different LLMs through lightweight adapters.
Intervention experiments suggest that the verbalizer is sensitive to compositional and signed relationships among SAE decoder directions.
Together, these results support internal representation verbalization as a trainable and partially reusable complement to methods that infer SAE feature meanings from activation examples, providing a more direct route from learned representations to feature explanations.


\section*{Limitations}

\paragraph{LLM and SAE Coverage.}
We primarily evaluate Gemma LLMs and Gemma Scope 2 SAEs, so generalization to other LLM families, SAE architectures, and representation spaces remains to be established.

\paragraph{Explanation Validation.}
Reference Agreement measures generated explanations' consistency with filtered reference explanations rather than the absolute correctness of the generated explanations.
The qualitative and intervention analyses provide complementary evidence, but do not systematically establish explanation correctness.

\paragraph{Run Variability.}
Each configuration is evaluated from a single run, so variability across random seeds is not measured.

\paragraph{Supervision Construction.}
Our current fine-tuning setup relies on high-quality feature--explanation supervision constructed from filtered Neuronpedia explanations, which is computationally expensive to produce.
This dependence is specific to our current supervision pipeline rather than inherent to \textsc{SAEVerbalizer}, as suitable supervision could also be constructed using alternative approaches, such as agentic or top-down methods.

\section*{Ethical Considerations}

\paragraph{Research Use and External Artifacts.}
SAEVerbalizer is intended as a research tool for interpreting learned representations.
We use the LLMs, SAEs, datasets, and online resources solely for academic research on LLM interpretability, consistent with their documented intended uses where specified and in accordance with their respective licenses or terms of use.

\paragraph{Reliability and Dual-Use Implications.}
Because SAEVerbalizer may inherit errors from the observational explanations used as supervision, its outputs should not be treated as verified descriptions of LLM behavior without independent validation.
More generally, improved access to internal feature meanings may support LLM auditing for safety and reliability, but may also enable more targeted manipulation of LLM representations.

\paragraph{Use of AI Assistants in Coding and Writing.}
AI assistants were used to support code development and language polishing.
All AI-assisted code and text were reviewed and revised as needed by the authors, who take full responsibility for the experiment implementation, analyses, and scientific claims presented in this work.

\bibliography{1-references}

\clearpage
\appendix
\section{Data Curation and Filtering Details}
\label{app:data_filtering}

This section describes the preliminary observations, two-stage filtering procedure, construction of the training and test splits, and prompts for the filtering judge.

\subsection{Preliminary Observations}
\label{app:data_preliminary}

Before constructing the verbalizer training data, we examined the reliability of Neuronpedia feature explanations~\cite{lin2023neuronpedia}.
We manually inspected 200 randomly sampled features from the \texttt{resid\_post/layer\_31\_width\_262k\_l0\_medium} checkpoint of \texttt{google/gemma-scope-2-27b-it}, a Gemma Scope 2 SAE defined on the layer-31 post-layer residual-stream representations of \texttt{gemma-3-27b-it}~\cite{gemmascope2_techreport}.
For 155 features (77.5\%), the top-activating text examples were semantically heterogeneous and did not reveal a sufficiently coherent activation pattern.
Among the remaining 45 features, the existing Neuronpedia explanation was judged accurate for 23 and inaccurate for 22; consequently, only 11.5\% of
the full sample exhibited both a coherent activation pattern and an accurate explanation.
These findings support explicit quality filtering before Neuronpedia explanations are adopted as training supervision or evaluation references.

We also observed that qualified feature--explanation pairs were more concentrated among lower-index features.
This tendency is qualitatively consistent with the Matryoshka organization of Gemma Scope 2 SAEs~\cite{gemmascope2_techreport} and motivates retaining a low-index test set alongside globally sampled test sets.

\subsection{Candidate Construction and Annotation}

Feature explanations and cached activation examples are obtained from Neuronpedia's \texttt{v1} bulk export archive~\cite{lin2023neuronpedia}.
They span multiple languages and heterogeneous conversational content; we apply no language-based filtering, and contributor demographic metadata are
unavailable.

For each SAE configuration, we construct a candidate pool from features with non-empty Neuronpedia explanations and rank their cached activation examples
by maximum activation.
We discard examples with invalid activation records, deduplicate rendered snippets after case-insensitive whitespace normalization, and retain up to 12 unique examples.

Each example is rendered as a local window extending up to 18 tokens on either side of the maximum-activation token.
The maximum-activation token is marked with \texttt{[[...]]}, while contiguous spans whose activations reach at least 60\% of the example-wise
maximum are marked with \texttt{<<...>>}.
Overlapping markers are nested as in \texttt{<<...[[token]]...>>}, and activated whitespace tokens are rendered as \texttt{[WHITESPACE]}.

\subsection{Two-Stage Filtering}

We use \texttt{Qwen3-30B-A3B-Instruct-2507}~\cite{qwen3technicalreport}
as the judge for both filtering stages.
Inference is run locally with \texttt{vLLM}~\cite{10.1145/3600006.3613165}
at temperature 0.

Stage 1 jointly presents the reference explanation and up to 12 marked activation examples and evaluates example-level matches, activation consistency, and explanation specificity.
Candidates pass when both scalar scores are at least 3 and match coverage is at least 0.75.

Candidates passing Stage 1 undergo a stricter Stage 2 assessment using the first eight examples, or all available examples when fewer than eight are retained.
Stage 2 evaluates activation consistency, monosemantic explanation specificity, and example-level matches.
The \textit{training qualification standard} requires both scalar scores to be at least 4 and match coverage to be at least 0.875; the \textit{gold qualification standard} raises both scalar thresholds to 5 while retaining the same coverage threshold.

As a sanity check, we manually inspected 50 randomly sampled retained pairs and found their activation examples generally coherent and their explanations
consistent with the marked patterns.

\subsection{Split Construction}

We construct the splits separately for each SAE configuration using fixed random seeds, first reserving 200 gold-qualified pairs from the low-index
region as the \textit{Low-Index Gold} (LIG) set.
From the remaining qualified pairs, we globally sample 1,000 gold-qualified pairs for the \textit{Global Gold} (GG) set and then 1,000 training-qualified pairs for the \textit{Global Train-Standard} (GTS) set.

After excluding all test pairs, we randomly order the remaining training-qualified pairs.
For SAEs with at least 48k remaining pairs, the first 48k form the largest training set, with the first 24k and 12k forming nested subsets.
For SAEs with fewer than 48k pairs, we use the largest available target size, either 24k or 12k, together with its corresponding nested subsets.

\subsection{Prompt Templates for the Filtering Judge}
\label{app:filter_prompts}

The exact prompts and user-message templates are presented below.

\subsubsection{Stage 1: Coarse Filtering Prompt}

\begin{promptbox}[Stage 1 System Prompt]
You are evaluating one sparse-autoencoder feature from its activation examples.

Notation:\\
- [[...]] is the maximum-activation token: the primary evidence.\\
- <<...>> is a region at or above 60\% of that maximum: supporting evidence.\\
- Text outside the marks is context, not the activation target.

Central rule:\\
Evaluate the localized property of the maximum-activation token. Do not reward explanations that merely describe the whole sentence, the broad topic, or a vague fact that any marked word has neighboring context.

Make three separate judgments in this order:

1. example\_matches\\
Judge each example independently before considering consistency across the set. Return 1 when the maximum-activation token and its local contextual role are directly covered by any specific part of the reference explanation. Otherwise return 0. If the explanation is a comma-separated list, matching one listed alternative is sufficient for that example. Supporting marked tokens need not literally appear in the explanation, but they must not contradict the match. A low overall consistency score must never force a direct individual match to 0. A condition, caveat, or topic mentioned elsewhere in the context does not make an unrelated marked token match; the claimed pattern must be localized at the marked position or in its relation to the preceding context. If the reference explanation is so broad that it would match almost any word in context, return 0 unless it names a concrete local role visible at the marked token.

2. activation\_consistency (1-5)\\
Ignore the reference explanation for this score. Compare what the maximum activation represents across examples: its word sense or its recurring semantic, syntactic, discourse, or structural role. Different surface words may be consistent only when that role clearly recurs. Similar sentence topics alone are not evidence of consistency. Use this calibration:\\
- 5: nearly every example has the same clear localized role, with no material counterexamples. Do not use 5 for a vague role such as "words with context", "sentence endings", "important words", or "related terms".\\
- 4: a large majority has that role, with only minor variation.\\
- 3: one role recurs, but the set has substantial mixed evidence.\\
- 2: only a minority shares a plausible role.\\
- 1: the marked positions are mostly unrelated.

3. explanation\_specificity (1-5)\\
Judge whether the explanation gives one precise, informative account of the activation pattern. Broad labels that can be retrofitted to unrelated contexts score low. A list of unrelated words is not a specific unified explanation. Use 5 for a precise near-complete account, 4 for a clear account with minor overbreadth, 3 for a useful but moderately broad account, 2 for a vague umbrella label, and 1 for an unrelated enumeration or unusably broad label. Specificity must be at most 2 for labels like "words with subsequent context", "sentence endings", "headings, descriptions, or reports", or broad lists of unrelated words, unless the marked tokens share a much more concrete local role explicitly named by the explanation.

Evidence constraints:\\
- Require evidence at the marked position together with preceding context. Do not match merely because unmarked nearby text discusses a related topic.\\
- Following text may clarify interpretation, but cannot cause an activation in a causal language model.\\
- [WHITESPACE], punctuation, and token fragments may be genuine structural evidence.\\
- Be conservative when evidence is ambiguous, but do not erase clear literal matches merely because other examples are inconsistent.\\
- If the maximum token is next to a structure, do not pretend the feature is the structure itself. For example, a digit before "))" is not a "closing parenthesis" feature unless the explanation says "token immediately before closing parentheses" or similar.\\
- A good feature may activate on different surface words, but then the shared role must be concrete and local, e.g. "CSS color property values" or "Dart constructor colon before super()", not merely "context words".

Return JSON only, with no explanation or additional keys:\\
\{"example\_matches": [0, 1], "activation\_consisten-\\cy": 1, "explanation\_specificity": 1\}
\end{promptbox}

\begin{promptbox}[Stage 1 User Prompt Template]
Feature ID: \{feature ID\}\\
Reference explanation: '\{reference explanation\}'\\

Number of activation examples: \(N\).
Return exactly \(N\) binary values in
\texttt{example\_matches}.\\

Activation examples:\\
Example 1: \{marked activation snippet\}\\
Example 2: \{marked activation snippet\}\\
\(\ldots\)\\
Example \(N\): \{marked activation snippet\}
\end{promptbox}

\subsubsection{Stage 2: Fine Filtering Prompt}

\begingroup
\tcbset{ignore nobreak=true}

\begin{promptbox}[Stage 2 System Prompt]
You are judging whether one sparse-autoencoder feature is high-quality and monosemantic.

You will see a reference explanation and activation examples. In each example:\\
- [[...]] marks the maximum-activation span. This is the primary evidence.\\
- <<...>> marks additional tokens above the activation threshold. These are supporting evidence.\\
- Unmarked text is context. Context can change the meaning or function of the marked span, but the evidence must stay localized around the marked span.

Your job is to judge exactly three things:

1. activation\_consistency (1-5)\\
Compare the context-conditioned meaning, linguistic function, discourse role, or structural role of the marked activation spans across examples.\\
- Give 5 when nearly all marked spans express the same clear local property, with no material counterexamples.\\
- Give 4 when a large majority express the same property with only minor variation.\\
- Give 3 when there is a recurring pattern but substantial mixed evidence.\\
- Give 2 when only a minority share a plausible pattern.\\
- Give 1 when the marked spans are mostly unrelated.

Different surface words may still be consistent if they have the same local meaning/function in context. Similar sentence topics are NOT enough.

2. explanation\_specificity\_monosemantic (1-5)\\
Judge whether the reference explanation gives one clear, specific, monosemantic account of the activation pattern.\\
- Give 5 for a precise near-complete explanation.\\
- Give 4 for a clear explanation with minor overbreadth.\\
- Give 3 for a useful but moderately broad explanation.\\
- Give 2 for a vague umbrella label.\\
- Give 1 for an unrelated list, grab bag, or unusably broad label.

Comma-separated explanations are acceptable only if the listed terms describe one unified local property. If the list combines different meanings/functions, score low even when each individual example matches one listed word.

3. matches\\
For each example independently, return 1 if the marked activation span's context-conditioned meaning/function is accurately covered by the reference explanation. Return 0 otherwise.

Important rules:\\
- Do not reward explanations that describe only the whole sentence topic.\\
- Do not reject punctuation, code, numbers, URLs, templates, or function words merely because they are structural. They can be excellent if the same specific role recurs.\\
- Do reject features that mix unrelated marked-span meanings/functions, even if the surrounding contexts look similar.\\
- Be conservative for vague explanations such as "things", "terms", "descriptions", "important words", "context words", or broad comma lists.

Return JSON only, with exactly these keys:\\
\{"activation\_consistency": 1-5, "explanation\_spec-\\ificity\_monosemantic": 1-5, "matches": [0 or 1 for each example], "reason": "short reason"\}
\end{promptbox}
\endgroup

\begin{promptbox}[Stage 2 User Prompt Template]
Feature ID: \{feature ID\}\\
Reference explanation: '\{reference explanation\}'\\

Number of examples: \(N\).
Return exactly \(N\) binary values in \texttt{matches}.\\

Example 1\\
anchor\_span: '\{maximum-activation token\}'\\
activated\_spans: [\{activated spans\}]\\
context: \{marked activation snippet\}\\

Example 2\\
anchor\_span: '\{maximum-activation token\}'\\
activated\_spans: [\{activated spans\}]\\
context: \{marked activation snippet\}\\

\(\ldots\)\\

Example \(N\)\\
anchor\_span: '\{maximum-activation token\}'\\
activated\_spans: [\{activated spans\}]\\
context: \{marked activation snippet\}
\end{promptbox}

\section{Verbalizer Implementation Details}
\label{app:verbalizer_implementation_details}
This section provides implementation details for the verbalizers,
complementing the injection interface and training objective defined in Sections~\ref{sec:verbalizer_design} and~\ref{sec:verbalizer_training}.

\subsection{Prompt and Injection Implementation}
\label{app:verbalizer_prompt}

We use the same prompt during training and inference.
It consists of a feature-agnostic instruction followed by the answer stem ``The target concept is:'':

\begin{promptbox}
You are an expert in concept interpretation.\\
I will inject an internal intervention into your hidden states.\\
Please complete the final sentence based on the intervention you experience.\\
The target concept is:
\end{promptbox}

The prompt is formatted using the chat template of the verbalizer backbone, with the generation cue appended.
The final four tokens of the formatted prompt constitute the default injection span.
During training, the reference explanation is appended and loss is computed only over its tokens; during inference, the verbalizer receives only the fixed
prompt and injected direction and generates autoregressively.

We implement the injection using a forward hook at the output of the selected Transformer layer.
Each instance in a batch receives its corresponding decoder direction, with norm matching computed independently for that instance.

\subsection{Training and Inference Configuration}
\label{app:common_verbalizer_settings}

All trained verbalizer configurations use Gemma Scope 2 checkpoints following the repository pattern
\path{google/gemma-scope-2-[scale]-it/resid_post/}
\path{layer_[L]_width_262k_l0_medium}, where the backbone scale and layer \(L\) are specified in
Section~\ref{sec:experiment_setup}.
The SAE-dictionary transfer experiment additionally uses
\path{google/gemma-scope-2-27b-it/resid_post/layer_16_width_65k_l0_medium}.
Table~\ref{tab:verbalizer_training_configuration} summarizes the remaining training and inference settings, where \(L_{\mathrm{tot}}\)
denotes the total number of Transformer layers.

\begin{table}[hbtp]
\centering
\small
\setlength{\tabcolsep}{5pt}
\begin{tabular}{ll}
\toprule
\textbf{Setting} & \textbf{Value} \\
\midrule
Injection operation &
Norm-matched additive \\
Injection strength &
\(\alpha=0.2\) \\
Frozen Transformer layers &
\(0,\ldots,L_{\mathrm{inj}}\) \\
Trainable Transformer layers &
\(L_{\mathrm{inj}}+1,\ldots,L_{\mathrm{tot}}-1\) \\
Other trainable modules &
Final norm and LM head \\
Other frozen modules &
\makecell[l]{Token embeddings\\
vision-related parameters} \\
Batch size &
\(4\) \\
Gradient accumulation &
None \\
Optimizer &
AdamW \\
Learning rate &
\makecell[l]{
\(5\times10^{-5}\) (1B and 4B)\\
\(1.5\times10^{-5}\) (27B)
} \\
Learning-rate schedule &
Constant \\
Warmup &
None \\
Weight decay &
\(0.01\) \\
Training epochs &
\(1\) \\
Numerical precision &
bfloat16 \\
Random seed &
\(42\) \\
\midrule
Decoding &
Greedy \\
Maximum new tokens &
\(15\) \\
\bottomrule
\end{tabular}
\caption{
Training and inference settings for the verbalizer configurations.
}
\label{tab:verbalizer_training_configuration}
\end{table}

\section{Adapter Implementation Details}
\label{app:adapter_details}
This section provides implementation details for the cross-LLM adapters.
Their architecture and training objective are defined in Sections~\ref{sec:adapter_design} and~\ref{sec:adapter_training}.

\paragraph{Representation-Pair Construction.}
We train separate adapters from layer 7 of
\texttt{gemma-3-1b-it} and layer 9 of \texttt{gemma-3-4b-it} to layer 16 of \texttt{gemma-3-27b-it}.
Because the models share the same tokenizer, each text is processed using identical input IDs and attention masks.
Sequences are appended with an end-of-sequence token and truncated or padded to 512 tokens.
Forward hooks collect the selected post-layer hidden states, and padding positions are removed to obtain token-aligned source--target representation pairs.

Both LLMs remain frozen and are run in bfloat16 during representation extraction.
Target representations are collected from the 27B verbalizer backbone.
Because the default verbalizer training freezes all Transformer layers through layer 16, the target representation space is unchanged in the final verbalizer.

\paragraph{Training Data and Optimization.}
We use the \texttt{sample-10BT} configuration of
\path{HuggingFaceFW/fineweb-edu}
(repository path \path{sample/10BT})
as the unlabeled corpus~\cite{NEURIPS2024_370df50c}.
FineWeb-Edu consists predominantly of English-language educational web text, and demographic metadata about the underlying contributors are unavailable.
A fixed representation validation set of 20,000 aligned non-padding representation pairs is constructed from the first 50 streaming examples.
To avoid overlap, training-data collection begins after the first 80 examples, and the remaining text stream is shuffled using a 5,000-example buffer.

Aligned representations are stored in bfloat16 in a CPU buffer containing up to one million representation pairs.
Each full buffer is randomly shuffled and divided into optimization batches of 8,192 representation pairs.
Table~\ref{tab:adapter_configuration} reports the remaining optimization settings.

\begin{table}[hbtp]
\centering
\small
\setlength{\tabcolsep}{4pt}
\begin{tabular}{@{}ll@{}}
\toprule
\textbf{Setting} & \textbf{Value} \\
\midrule
Adapter &
Affine layer \\
Initialization &
Zero weight and bias \\
Objective &
MSE \\
Optimizer &
AdamW \\
Learning rate &
\(1\times10^{-4}\) \\
Weight decay &
\(0.01\) \\
Learning-rate schedule &
200-step warmup, then constant \\
Training steps &
5,000 \\
Adapter precision &
float32 \\
Random seed &
42 \\
\bottomrule
\end{tabular}
\caption{
Optimization settings shared by both cross-LLM adapters.
}
\label{tab:adapter_configuration}
\end{table}

\paragraph{Validation and Checkpoint Selection.}
On the fixed validation set, we monitor MSE and mean token-level cosine similarity.
These metrics serve only as representation-alignment diagnostics.

Every 1,000 steps, we evaluate a candidate checkpoint by mapping the source-SAE decoder directions into the default \texttt{27B-L16} verbalizer and computing RA on the corresponding source SAE's 200-feature LIG set.
This selects step 3,000 for the 1B-to-27B adapter and step 4,000 for the 4B-to-27B adapter; GTS and GG are not used for checkpoint selection.

\paragraph{Decoder-Direction Mapping and Inference.}
At inference, each source SAE decoder direction is mapped in float32 using only the adapter weight matrix,
\(\tilde{\mathbf{v}}^{(t)}_f=W\mathbf{v}^{(s)}_f\), without the affine bias.
The mapped direction is then supplied to the fixed
\texttt{27B-L16} verbalizer through the same norm-matched injection, prompting, and generation interface used for native decoder directions.

\section{Evaluation Protocol Details} \label{app:evaluation_protocol} 
This section specifies explanation generation and the LLM-judge protocol for Reference Agreement.

\paragraph{Explanation Generation.} 
For each test feature, we generate one explanation using the prompt, injection, and decoding configuration described in Appendix~\ref{app:verbalizer_implementation_details}, without providing the reference explanation or activation examples.

\paragraph{Reference Agreement.} 
We compare each generated explanation with its reference using an LLM evaluation judge.
The judge is
\texttt{Qwen3-30B-\allowbreak A3B-Instruct-\allowbreak 2507}~\cite{qwen3technicalreport},
run locally with \texttt{vLLM}~\cite{10.1145/3600006.3613165} in bfloat16
at temperature 0 with a maximum of three new tokens.
The reference and generated explanations are presented as Concept A and Concept B, respectively.
The judge returns \texttt{YES} for synonymous, strongly overlapping, subset--superset, or directly associated concepts, and \texttt{NO} otherwise.
Reference Agreement (RA) is the percentage of generated explanations judged \texttt{YES}.

\paragraph{Prompt Template for the Evaluation Judge.} 
The exact prompt template is presented below.
\begin{promptbox}[Prompt for Reference Agreement Evaluation] \textbf{System:}\\ 
You are an expert semantic judge for concept extraction. Compare Concept B, the model prediction, with Concept A, the reference. 
Answer YES if they are synonymous, strongly overlapping, one is a subset or superset of the other, or they have a strong direct logical/contextual association. 
Repetition and harmless wording differences must still be YES. 
Answer NO only when the concepts are meaningfully unrelated.
Output exactly YES or NO.
\\[4pt] 
\textbf{User:}\\ 
Concept A: '\{reference explanation\}'\\
Concept B: '\{generated explanation\}'\\ 
Is Concept B a semantic hit for Concept A? 
\end{promptbox}

\section{Additional Verbalizer Analyses}
\label{app:additional_verbalizer_analyses}

This section reports the full robustness results summarized in
Section~\ref{sec:ablations} and additional qualitative comparisons and joint-injection cases from Section~\ref{sec:case_studies}.

\subsection{Prompt and Injection Robustness}
\label{app:prompt_injection_robustness}

All experiments use the default 48k \texttt{27B-L16} configuration, vary one factor at a time, and use the same variant during training and inference.

\paragraph{Prompt Variants.}
Starting from the default prompt reported in Appendix~\ref{app:verbalizer_prompt}, we evaluate four semantically similar variants.
The \textit{SAE-aware instruction} replaces the feature-agnostic instruction with: 
\begin{promptbox} 
You are an expert in SAE feature interpretation.\\ 
I will inject an internal SAE feature vector into your hidden states.\\ 
Please complete the final sentence by naming the target concept represented by the injected feature. 
\end{promptbox}

The \textit{specific-label instruction} inserts
``Answer with a specific label.'' immediately before the default answer stem, ``The target concept is:''.
The other two variants replace this stem with
``The intervention represents:'' and
``The injected intervention reveals:'', which we refer to as the \textit{representation variant} and \textit{revelation variant}, respectively.

\begin{table}[hbtp]
\centering
\scalebox{0.89}{
\begin{tabular}{lrrr}
\toprule
\textbf{Prompt variant} &
\multicolumn{1}{c}{\textbf{GTS}} &
\multicolumn{1}{c}{\textbf{LIG}} &
\multicolumn{1}{c}{\textbf{GG}} \\
\midrule
Default prompt              & \(52.3\) & \(80.5\) & \(56.1\) \\
SAE-aware instruction       & \(53.2\) & \(79.5\) & \(55.0\) \\
Specific-label instruction  & \(52.2\) & \(78.0\) & \(54.9\) \\
Representation variant      & \(53.7\) & \(81.5\) & \(54.4\) \\
Revelation variant              & \(53.1\) & \(80.0\) & \(55.5\) \\
\bottomrule
\end{tabular}
}
\caption{
RA (\%) under semantically similar prompt variants.
}
\label{tab:prompt_robustness}
\end{table}

\begin{table*}[t]
\centering
\small
\setlength{\tabcolsep}{4pt}
\renewcommand{\arraystretch}{1.05}

\begin{tabularx}{\textwidth}{
@{}
>{\raggedright\arraybackslash}m{0.13\textwidth}
>{\raggedright\arraybackslash}p{0.065\textwidth}
>{\raggedright\arraybackslash}p{0.18\textwidth}
>{\raggedright\arraybackslash}p{0.14\textwidth}
X
@{}
}
\toprule
\textbf{Pattern} &
\textbf{Feature} &
\textbf{Neuronpedia} &
\textbf{Verbalizer} &
\textbf{Activation Evidence} \\
\midrule

\multirow[c]{2}{=}[-4ex]{\textbf{Recurring lexical form}} &
\#12488 &
introduces contingencies &
sometimes &
The maximally activated token is consistently \textit{sometimes} across
varied topics and sentence positions. \\

\arrayrulecolor{black!20}
\cmidrule(lr){2-5}
\arrayrulecolor{black}

&
\#64782 &
describing bad or unpleasant things &
nasty &
The maximally activated token is consistently \textit{nasty} across
varied entities and contexts, such as people, infections, substances,
and software. \\

\arrayrulecolor{black!20}
\specialrule{0.4pt}{2pt}{2pt}
\arrayrulecolor{black}

\multirow[c]{2}{=}[-2.0ex]{\textbf{Localized structural pattern}} &
\#7807 &
just a basic descriptor &
mere followed by a qualifier &
The activation examples are dominated by attributive \textit{mere}
before a following noun phrase. \\

\arrayrulecolor{black!20}
\cmidrule(lr){2-5}
\arrayrulecolor{black}

&
\#83752 &
guessing or asking questions &
guess what &
The activation examples predominantly contain \textit{guess what}
and related attention-directing question frames. \\

\arrayrulecolor{black!20}
\specialrule{0.4pt}{2pt}{2pt}
\arrayrulecolor{black}

\multicolumn{1}{
@{}>{\raggedright\arraybackslash}p{0.13\textwidth}
}{
\textbf{Contextual semantic abstraction}
} &
\#98611 &
utmost care, respect, or seriousness &
utmost importance &
The maximally activated token is consistently \textit{utmost}, followed
by varied abstract nouns such as \textit{respect}, \textit{seriousness},
\textit{precision}, \textit{care}, and \textit{responsibility}. \\

\bottomrule
\end{tabularx}

\caption{
Additional qualitative comparisons between verbalizer-generated and
Neuronpedia explanations.
}
\label{tab:additional_qualitative_comparisons}
\end{table*}

\paragraph{Injection-Span Variants.}
Under the default prompt, the Gemma chat template appends the generation cue
\texttt{<end\_of\_turn>\textbackslash n <start\_of\_turn> model \textbackslash n}.
In Table~\ref{tab:span_robustness}, we compare the default injection into its final four tokens with injection into the answer stem, the span from the stem through the cue, and the full cue.

\begin{table}[hbtp]
\centering
\scalebox{0.89}{
\begin{tabular}{lrrr}
\toprule
\textbf{Injection span} &
\textbf{GTS} &
\textbf{LIG} &
\textbf{GG} \\
\midrule
Final four tokens
& $52.3$ & $80.5$ & $56.1$ \\

Answer stem
& $53.0$ & $79.0$ & $54.6$ \\

Stem through cue
& $52.9$ & $80.5$ & $53.1$ \\

Full generation cue
& $54.1$ & $79.0$ & $54.1$ \\
\bottomrule
\end{tabular}
}
\caption{
RA (\%) under different injection spans.
}
\label{tab:span_robustness}
\end{table}

\paragraph{Injection Mode.}
In Table~\ref{tab:injection_mode_robustness}, we compare the default norm-matched additive injection with the
interpolative variant
\(\mathbf{h}'_{b,s}
=(1-\alpha)\mathbf{h}_{b,s}
+\alpha\bar{n}_b\hat{\mathbf{v}}_{f_b}\)
at \(\alpha\in\{0.1,0.2,0.3\}\).

\begin{table}[hbtp]
\centering
\scalebox{0.89}{
\begin{tabular}{lrrrr}
\toprule
\textbf{Mode} &
\multicolumn{1}{c}{\(\boldsymbol{\alpha}\)} &
\multicolumn{1}{c}{\textbf{GTS}} &
\multicolumn{1}{c}{\textbf{LIG}} &
\multicolumn{1}{c}{\textbf{GG}} \\
\midrule
Additive
& \(0.10\) & \(50.6\) & \(78.5\) & \(54.2\) \\
& \(0.20\) & \(52.3\) & \(80.5\) & \(56.1\) \\
& \(0.30\) & \(52.8\) & \(81.0\) & \(54.8\) \\
\midrule
Interpolation
& \(0.10\) & \(52.1\) & \(81.0\) & \(52.9\) \\
& \(0.20\) & \(51.2\) & \(80.0\) & \(54.1\) \\
& \(0.30\) & \(53.1\) & \(82.0\) & \(54.9\) \\
\bottomrule
\end{tabular}
}
\caption{
RA (\%) under additive and interpolative injection modes at different
injection strengths.
}
\label{tab:injection_mode_robustness}
\end{table}

Across the three comparisons, RA varies only modestly under changes to prompt wording, injection span, and injection mode.

\subsection{Additional Case Studies}
\label{app:additional_case_studies}

\paragraph{Qualitative Comparisons.} Table~\ref{tab:additional_qualitative_comparisons} extends Table~\ref{tab:qualitative_explanation_comparison} with additional cases involving recurring lexical forms, localized structural patterns, and different semantic abstractions.

\paragraph{Joint-Injection Cases.} \label{app:additional_joint_injection} Table~\ref{tab:additional_feature_composition} extends Table~\ref{tab:feature_composition_cases} with additional feature pairs evaluated under the same joint-injection protocol. The two directions receive equal coefficients whose sum equals the default single-feature injection strength, \(\alpha=0.2\).

\begin{table}[hbtp] 
\centering 
\small 
\setlength{\tabcolsep}{3pt} 
\renewcommand{\arraystretch}{1.12} 
\begin{tabular}{ 
@{} 
l 
>{\raggedright\arraybackslash}p{0.43\columnwidth} 
>{\raggedright\arraybackslash}m{0.31\columnwidth} 
@{} 
} 
\toprule 
\textbf{Feature} & 
\textbf{Standard} & 
\textbf{Joint} \\ 
\midrule 
\#95421 & firmware updates and security & \multirow{2}{=}[-1.5ex]{software version numbers} \\ 
\#16153 & version numbers & \\ 
\arrayrulecolor{black!20} 
\specialrule{0.4pt}{2pt}{2pt} 
\arrayrulecolor{black} 
\#8820 & symptoms of conditions & \multirow{2}{=}[-1.5ex]{assessment and symptom} \\ 
\#7782 & assessment and assessment tools & \\ \arrayrulecolor{black!20} 
\specialrule{0.4pt}{2pt}{2pt} 
\arrayrulecolor{black}
\#1586 & coffee and its contexts & \multirow{2}{=}{morning coffee} \\ 
\#6502 & morning time & \\ 
\arrayrulecolor{black!20} 
\specialrule{0.4pt}{2pt}{2pt} 
\arrayrulecolor{black} 
\#13401 & fragrant scents and aromas & \multirow{2}{=}{coffee and its aroma} \\ 
\#1586 & coffee and its contexts & \\ 
\bottomrule 
\end{tabular} 
\caption{Additional examples of joint feature injection under the same configuration as Table~\ref{tab:feature_composition_cases}.} \label{tab:additional_feature_composition} 
\end{table}

\section{Computational Details}
\label{app:computational_details}

\paragraph{Software.}
Verbalizer and adapter training and evaluation use the Hugging Face Transformers implementations of the Gemma backbones~\cite{wolf-etal-2020-transformers}; filtering and evaluation judges use \texttt{vLLM}~\cite{10.1145/3600006.3613165} with
\texttt{Qwen3-30B-A3B-Instruct-2507}.

\paragraph{Computational Resources.}
Our experiments use Gemma 3 backbones at the 1B, 4B, and 27B parameter scales.
On NVIDIA A100 80GB GPUs, the largest 27B verbalizer runs take approximately two hours on four GPUs, while 1B and 4B runs take approximately one hour on one GPU.
Adapter training uses two NVIDIA A100 80GB GPUs, with the source and target backbones computing representations concurrently.

\paragraph{Run Reporting.}
Each trained verbalizer or adapter configuration is run once, and the reported quantitative results are not averaged across random seeds.

\end{document}